\documentclass{article}
\usepackage{arxiv}

\usepackage{mathtools} 

\usepackage{booktabs}
\usepackage{array,multirow}
\usepackage{rotating}
\usepackage{makecell}
\usepackage{rotating}
\usepackage{array}
\usepackage{float}
\usepackage{physics}
\usepackage{algorithm}
\usepackage{algpseudocode}
\usepackage{enumitem}
\usepackage[font={small,it}]{caption}
\usepackage{hyperref}
\usepackage{microtype}

\usepackage{amssymb}
\usepackage{amsmath}
\usepackage{mathdots}
\usepackage{mathtools}
\usepackage{multicol}

\usepackage{tensor}

\usepackage{urwchancal}
\DeclareFontFamily{OT1}{pzc}{}
\DeclareFontShape{OT1}{pzc}{m}{it}{<-> s * [1.10] pzcmi7t}{}
\DeclareMathAlphabet{\mathpzc}{OT1}{pzc}{m}{it}

\usepackage{graphicx}
\usepackage{ifpdf}

\ifpdf
\usepackage{epstopdf}
\PrependGraphicsExtensions{.svg}
\fi

\usepackage{booktabs}
\usepackage{array,multirow}
\usepackage{rotating}
\usepackage{makecell}
\usepackage{rotating}
\usepackage{array}
\usepackage{float}
\usepackage{physics}
\usepackage{algorithm}
\usepackage{algpseudocode}
\usepackage{bbding}
\usepackage{amsmath}
\usepackage{bm}

\usepackage{wrapfig}

\newcommand{\etal}{\textit{et al}. }

\newcommand{\eg}{\textit{e}.\textit{g}., }

\usepackage{adjustbox}
\usepackage{multirow}
\usepackage[dvipsnames]{xcolor}
\usepackage[normalem]{ulem}

\usepackage{adjustbox}
\usepackage{multirow}
\usepackage[dvipsnames]{xcolor}
\usepackage[normalem]{ulem}
\usepackage{xspace}

\makeatletter
\DeclareRobustCommand\onedot{\futurelet\@let@token\@onedot}
\def\@onedot{\ifx\@let@token.\else.\null\fi\xspace}

\def\etc{\emph{etc}\onedot}

\usepackage{array}
\newcolumntype{?}[1]{!{\vrule width #1}}

\usepackage{booktabs,tabularx}
\definecolor{pink}{rgb}{0.858, 0.188, 0.478}

\newcommand{\publicationdetails}
{Wang, Z., Ng, Y., Henderson, J., Mahony R. (2022), ``Smart Visual Beacons with Asynchronous Optical Communications using Event Cameras", \textit{published in IEEE International Conference on Intelligent Robots and Systems (IROS).} 
\copyrightNoticeIEEE{2022}	
	}
\newcommand{\publicationversion}
{Author Accepted Version}

\title{\LARGE \bf
	Smart Visual Beacons with Asynchronous Optical Communications using Event Cameras}

\author{
	\href{https://orcid.org/0000-0003-0815-1287}{\includegraphics[scale=0.06]{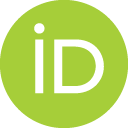}\hspace{1mm}
		Ziwei Wang}
	\\
	Systems Theory and Robotics Group \\
	Australian National University \\
	ACT, 2601, Australia \\
	\texttt{ziwei.wang1@anu.edu.au} \\
	\\
	\And
	\And
	\href{https://orcid.org/0000-0002-7764-298X}{\includegraphics[scale=0.06]{orcid.png}\hspace{1mm}
		Yonhon Ng}
	\\
	Systems Theory and Robotics Group \\
	Australian National University \\
	ACT, 2601, Australia \\
	\texttt{yonhon.ng@anu.edu.au} \\
	\\
	\And
	\href{https://orcid.org/0000-0003-1506-0316}{\includegraphics[scale=0.06]{orcid.png}\hspace{1mm}
		Jack Henderson}
	\\
	Systems Theory and Robotics Group \\
	Australian National University \\
	ACT, 2601, Australia \\
	\texttt{jack.henderson@anu.edu.au} \\
	\And	\href{https://orcid.org/0000-0002-7803-2868}{\includegraphics[scale=0.06]{orcid.png}\hspace{1mm}
		Robert Mahony}
	\\
	Systems Theory and Robotics Group \\
	Australian National University \\
	ACT, 2601, Australia \\
	\texttt{robert.mahony@anu.edu.au} \\
}

\begin{document}

\maketitle
\thispagestyle{empty}
\pagestyle{empty}


\begin{abstract}
Event cameras are bio-inspired dynamic vision sensors that respond to changes in image intensity with a high temporal resolution, high dynamic range and low latency.
These sensor characteristics are ideally suited to enable visual target tracking in concert with a broadcast visual communication channel for smart visual beacons with applications in distributed robotics.
Visual beacons can be constructed by high-frequency modulation of Light Emitting Diodes (LEDs) such as vehicle headlights, Internet of Things (IoT) LEDs, smart building lights, \etc, that are already present in many real-world scenarios.
The high temporal resolution characteristic of the event cameras allows them to capture visual signals at far higher data rates compared to classical frame-based cameras.
In this paper, we propose a novel smart visual beacon architecture with both LED modulation and event camera demodulation algorithms.
We quantitatively evaluate the relationship between LED transmission rate, communication distance and the message transmission accuracy for the smart visual beacon communication system that we prototyped.
The proposed method achieves up to 4 kbps in an indoor environment and lossless transmission over a distance of 100 meters, at a transmission rate of 500 bps, in full sunlight, demonstrating the potential of the technology in an outdoor environment.
\end{abstract}

\centerline{
	\noindent \tt\small Project URL: \href{https://github.com/ziweiWWANG/Event-Beacon-Communication.git}{{\color{pink}\texttt{https://github.com/ziweiWWANG/Event-Beacon-Communication.git}}}
}

\section{INTRODUCTION}
\begin{figure}[t!]
	\centering
	\begin{tabular}{cc}
		\\	\includegraphics[width=0.48\linewidth]{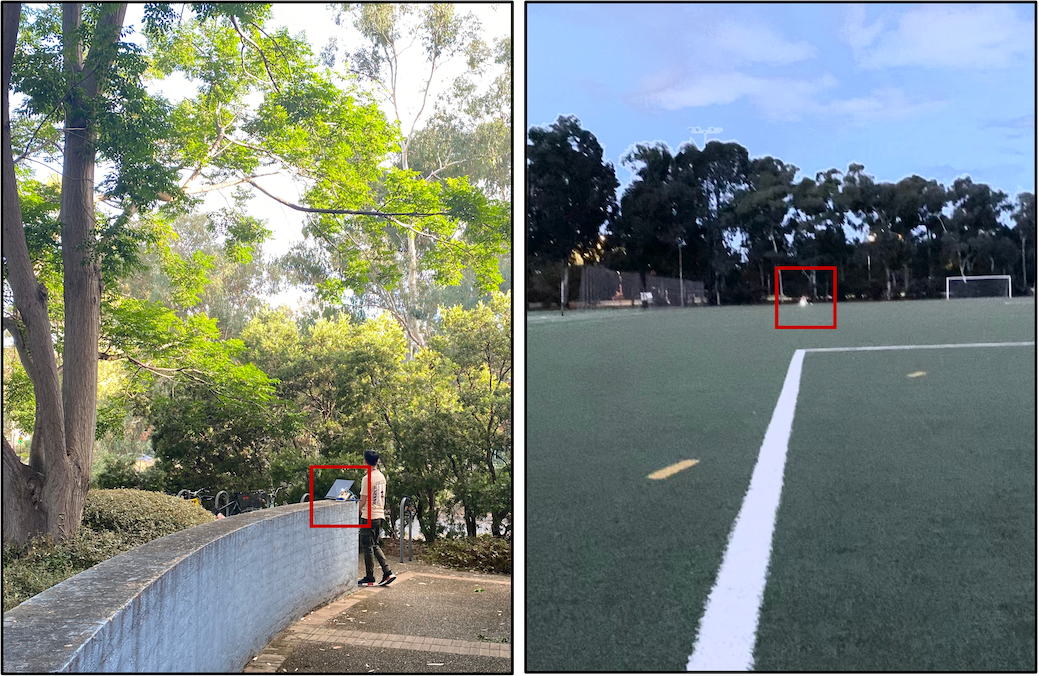}  &
		\includegraphics[width=0.36\linewidth]{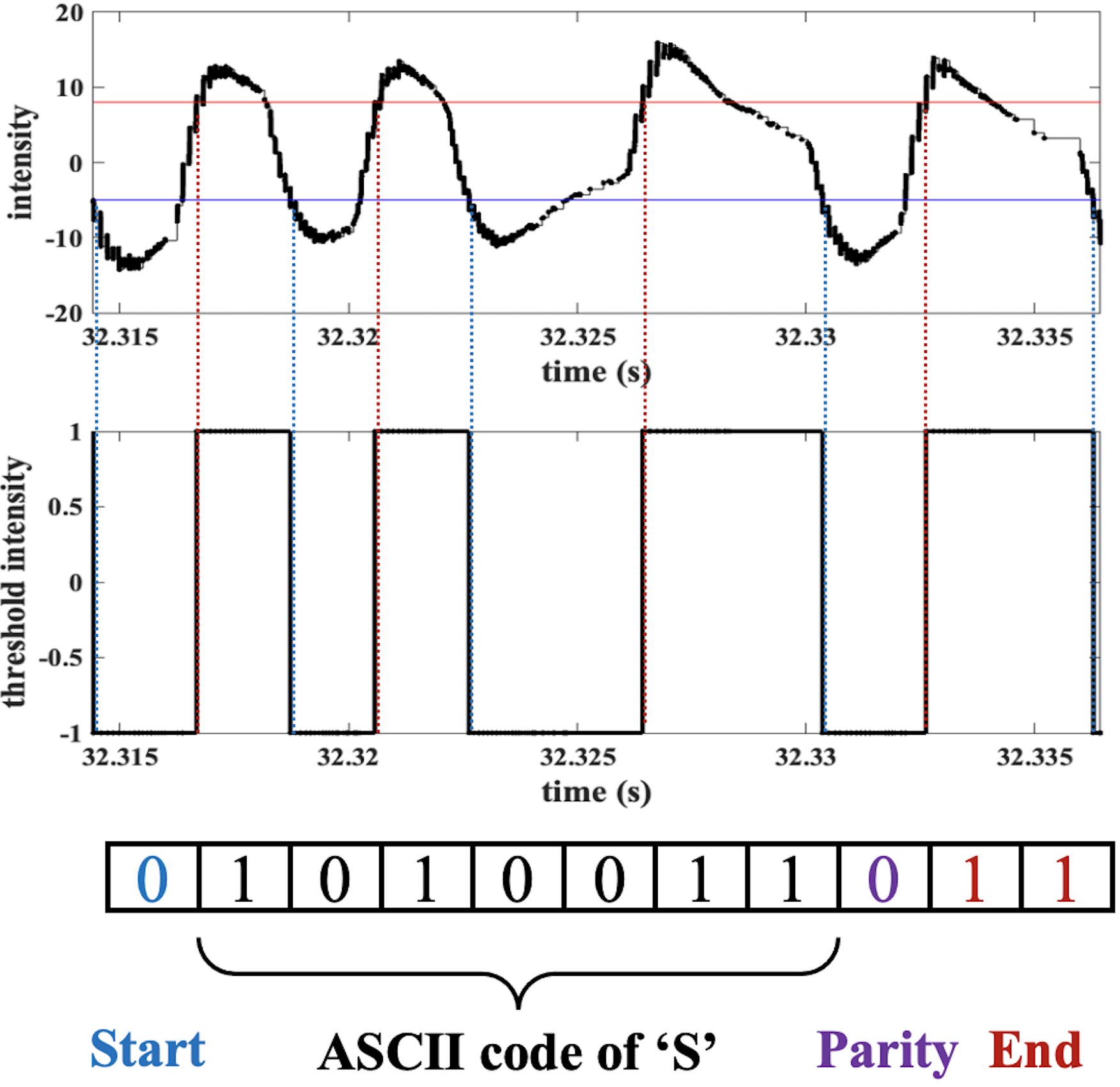} \\
		(a) Outdoor Daylight Experimental Environments
		 &
		(b) Example of 100-meter Error-Free Communication Data\\
	\end{tabular}
	\caption{		
		Outdoor Communication Using Our Smart Visual Beacon. (a) shows two outdoor daylight environments where we evaluate long-distance communication, at 7 m on the left, and 100 m distance on the right.
		The smart visual beacons are shown in red squares.
		The first scene shows that the complex background does not affect communication, and the second demonstrates 100-meter long-distance communication.
		The first row in (b) plots the asynchronous high-pass filtered~\cite{scheerlinck2018continuous} intensity received for a portion of the 100-meter error-free communication, and the red and blue horizontal lines represent the high and low-level thresholds respectively.
		The second row in (b) shows the binarised intensity.
		The data packet encoded was the letter ``S" (1010011) with the start, parity and end bits.
	}\label{fig:outdoor}
\end{figure}
Visual tracking of stationary and moving beacons or targets is an enabling technology in a wide range of robotics applications including logistics systems, automotive leader-follower systems, distributed robotics, \etc.
In most existing systems, the beacons are either distinguished by their geometric forms, such as ArUco markers, colour patterns or bar codes, or are a light source, visible or ultraviolet, distinguished by their colour or by a simple low-frequency flashing signal.
Such beacons may be recognised by a robot, however, they provide no additional communication capability.
Overcoming this limitation is particularly important in distributed robotics applications such as cooperative driving, smart cities, formation or swarm robotics, \etc, where it is desirable to communicate information to neighbouring agents, for example sharing its estimated state rather than simply broadcasting its identity.
This situation becomes even more important where traditional radio-frequency (RF) communication is unviable, due to high interference, background noise, bandwidth saturation, underwater applications, or intentional signal jamming \etc.
Optical camera communication (OCC) is a popular option for an alternative short-range wireless communication in line-of-sight conditions.
Low weight visual optical channels are relatively easy and cheap to be integrated into existing robotic systems~\cite{joubert2019characterization}.
Existing optical camera communication systems typically rely on traditional CMOS or CCD cameras to capture modulated blinking LED transmitters.
However, with off-the-shelf cameras, the data rates achieved by OCC systems are poor, typically in the range of 100 to 250 bits per second (bps) at best~\cite{saeed2019optical}.
Higher data rates can be achieved with more advanced high-speed cameras, or through complex transmitter and receiver systems that allow image sensors to capture down-sampled high-frequency information.
Such systems, however, require significant computational power or dedicated physical infrastructure.
Short exposure times in typical high speed cameras also require brighter LED transmitters leading to high power requirements on the beacon.

In contrast to traditional frame-based cameras that accumulate irradiance for all pixels within the exposure time, event cameras continuously monitor the logarithmic light intensity over each pixel independently and only generate events at pixels where brightness changes cross a pre-defined threshold.
This architecture results in low latency (milliseconds), high-dynamic range (greater than 120 dB), and lower power operation.
These characteristics make event cameras ideal for integration into existing sensor suites~\cite{wang2021stereo}, capable of detecting high-frequency modulated light sources generated by a smart beacon for high-speed, low-latency communication for autonomous robots.

\begin{figure*}[t!]
	\centering
	\begin{tabular}{c}
		\\	\includegraphics[width=1\linewidth]{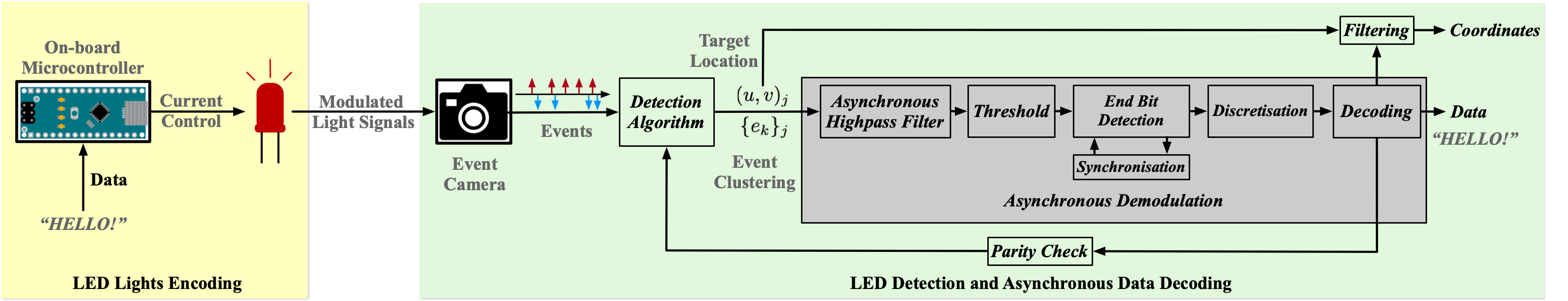}
		\\
	\end{tabular}
	\caption{
		Asynchronous Event-based Smart Visual Beacon Communication System Pipeline.
	}\label{fig:pipeline}
\end{figure*}

In this paper, we propose a smart visual beacon system that includes a broadcast optical camera communication channel.
The system exploits the particular characteristics of the event camera to obtain a high-frequency visual intensity signal from a visual beacon and a robust demodulation algorithm to decode the transmitted data.
Since event cameras only capture pixels with changes in brightness, a flickering light source, such as a visual beacon, is easily identified as it generates a large number of events per second compared to other signals in the scene~\cite{wang22icra}.
The proposed methodology exploits this property to track visual targets and then passes the associated signal into a demodulation algorithm that reconstructs the transmitted signal asynchronously using high-pass filter~\cite{scheerlinck2018continuous}.
Then the method applies thresholds to extract the binary bitstream, synchronises with the start bit automatically, and decodes the transmitted message.
To demonstrate the system, we transmit messages using the ASCII code.
The 7-bit ASCII code is augmented with a parity bit and the resulting 8-bit packet is encoded using the Universal Asynchronous Receiver-Transmitter (UART) protocol.
We emphasise that these choices are simply a first demonstration of capability and more sophisticated encoding schemes are likely to yield better results.
The resulting bitstream is transmitted using on and off modulation of an LED light.
For a typical data stream, the LED is on for more than 50\% of the time and, for high-frequency transmission, the resulting light source appears as a normal constant LED light to the human eyes.
The bit rate of the modulated signal can be pushed to over 4 kbps when the LED is sufficiently bright.
For higher bit rates, refractory period limitations and other nonlinearities of the event camera begin to limit data quality~\cite{Yang15ssc,Wang19acra,wang2021asynchronous}.
Similarly, we have demonstrated error-free transmission at 500 bps in full daylight over 100 m using a bright LED.
The LED we used is small in size and not brighter than a typical car headlight or bright torch.
It could easily be integrated into a small-scale robotic system such as a quadrotor.

In summary:
\begin{itemize}
	\item We propose a smart visual beacon system that integrates visual beacon tracking with a broadcast optical camera communication and demonstrate its capability for high data communication in real-world outdoor scenarios.
	
	\item We demonstrate the event-based smart visual beacon optical communication with real experiments and quantitatively analyse the performance at various bit rates and distances.
\end{itemize}

\section{RELATED WORK}
Recognising the advantages of high temporal resolution and low-latency characteristics of event cameras in detecting fast blinking LED markers, some event-LED-based pose estimation, feature tracking and communication algorithms are developed.
Censi~\etal~\cite{censi2013low} proposed a pose tracking algorithm using four infrared LED markers with different blinking frequencies which are distinguishable by event cameras.
They accumulated triggered events in short time windows and tracked the pose of the drone attached with LED markers by particle filter.
Instead of the traditional CMOS cameras, Chen~\etal~\cite{chen2020novel} used an event camera in their visible light positioning system with multi-LEDs, which largely reduced the computational cost and latency.
They estimated different frequencies of LED markers based on the time intervals between positive-negative transitions at a pixel and then tracked LED markers with their GM-PHD filter, resulting in an accuracy of 3cm when the distance between LEDs and camera is less than 1m.
Joubert~\etal~\cite{joubert2019characterization} characterised event camera LED setup for event-based imagers applied to modulated light signal detection.
Some preliminary results were also obtained by Perez-Ramire~\etal~\cite{perez2019optical} where events were first accumulated into synchronous frames. They then proposed a start frame delimiter (SFD) and symbol design derived from the Composite Waveform, and showed some Bit Error Rate (BER) results for transmission of 2 bytes packets.

However, the potential of the combined optical tracking and communication using event cameras has not been well-studied since most of the recent works still focus on traditional CMOS cameras.
Many algorithms are developed in the pose estimation and close-range tracking~\cite{faessler2014monocular}, with application in multi-agent unmanned aerial vehicles~\cite{dias2016board,walter2018mutual,walter2018fast} and autonomous underwater vehicles system~\cite{bosch2016close,gracias2015pose}.
Many standardised methods for LED light modulation~\cite{saeed2019optical} are developed for optical camera communication (OCC)~\cite{saeed2019optical},
such as under-sampled frequency shift on-off keying (UFSOOK)~\cite{roberts2013undersampled},
camera multiple frequency shift keying (CM-FSK)~\cite{rajagopal2014visual,lee2015rollinglight},
pulse width modulation/pulse position modulation (PWM/PPM)~\cite{aoyama2015line,aoyama2015visible},
spatial-2-phase-shift keying (S2-PSK)~\cite{nguyen2016region},
camera on-off keying (C-OOK)~\cite{nguyen2016high}, \etc.
There are also many different well-established methods and standards (\eg IEEE.802.15.7) for short-range optical wireless communications~\cite{saeed2019optical}.
Usually, to capture the high-frequency LED blinking signal, a high-speed camera is required, such as 1000 fps in~\cite{roberts2014automotive,ebihara2015layered}, or using a low frame rate CMOS camera with the selective capturing strategy~\cite{teli2018high}, under-sampled modulation~\cite{luo2014undersampled} or custom-built system~\cite{oike2003smart,imai2016performance}.

\section{EVENT-BASED SMART VISUAL BEACON COMMUNICATION SYSTEM}
Our smart visual beacons communication system consists of two main steps: LED signal encoding and asynchronous data decoding (shown in Figure~\ref{fig:pipeline}).
The encoding and decoding are done on separate systems, \eg encoding on a beacon base station and decoding on a mobile robot or vice versa.

\subsection{LED lights encoding}
\begin{figure}[t!]
	\centering
		\begin{tabular}{c}
			\\	\includegraphics[width=0.7\linewidth]{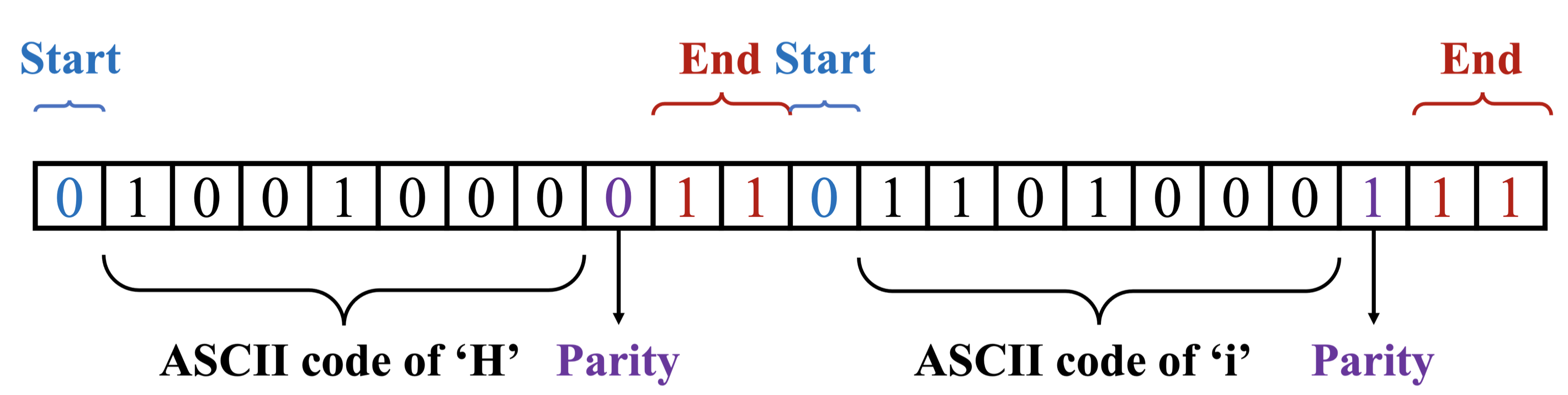}
		\end{tabular}
	\caption{
		The proposed LED Lights Encoding Method. Each character in the message is converted to a 7-bit binary ASCII code and encoded with the standard Universal Asynchronous Receiver-Transmitter (UART) protocol with a $\{0\}$ as start bit and $\{1\;1\}$ as end bit.
		A parity bit is appended to the original data packet as a simple and effective way to detect transmission errors.
	}\label{fig:encoding}
\end{figure}

The transmitted message is encoded into the standard ASCII code in which each character is encoded into a 7-bit binary code.
The message characters are then encoded into data packets using the UART protocol with a parity bit, a start bit, and two end bits (see Figure~\ref{fig:encoding}).
The signal is encoded by an Arduino Nano microcontroller which generates the modulated voltage signal to control the fast blinking of LED markers.
To detect the onboard LED marker easily, we set the voltage high as default to have the LED marker turned on when no message is being transmitted.
The parity bit plays a key role in the system in providing feedback to the detection algorithm and identifying the beacon signals from non-beacon ``noise'' signals in the scene (see~Figure~\ref{fig:pipeline} and Section~\ref{sec:Detection algorithm}).
We use a single zero as the start bit and two ones as the end bits for each 8-bit data packet to ensure each code is separated by a $\{1\;1\;0\}$ binary pattern.
The $\{1\;1\;0\}$ LED blinking pattern generates a strong falling edge at the end of each transmitted character in the continuous-time intensity reconstruction from event data and provides a clean signal used for synchronisation.
More details are covered in \S \ref{sec:End bit detection and synchronisation}.

\subsection{LED detection and asynchronous data decoding}
\subsubsection{Optical communication}
We use an event camera to capture the modulated LED blinking signal.
Although event cameras have high temporal resolution and low latency, they are still subjected to refractory period limitations and nonlinear effects~\cite{Wang19acra,wang2021asynchronous}.
Communication quality decreases when the transmission rate is pushed to the hardware limit. 
We will evaluate this in the experiment section (see Figure~\ref{fig:indoor}).

\subsubsection{Detection algorithm}\label{sec:Detection algorithm}
The LED detection module detects the LED transmitter pixels using conventional blob detection and tracking method.
A sliding window of events are used to compute a pixel-based index,
\[
I_{(u,v)} = \frac{\# \texttt{events}}{B + |\Sigma \sigma_k|},
\]
comprising the number of events divided by the absolute value of the sum of the polarities $\sigma_k$ of each individual event plus an offset $B > 0$, to ensure the result is always finite.
High indices indicate a flickering light source with many events evenly balanced between positive and negative polarities, typical of a communication channel.
A threshold is chosen to remove pixels with low indices and image morphology is applied to the resulting binary image to close holes within blobs and make the detection and tracking more robust.
Blob tracking is based on the nearest neighbour blob association.
Noting that the blob estimates associated with a given beacon will be overlapping as the sliding window moves through the data.
A unique label $j$ is associated with each tracked blob and the bearing of the blob is computed from the centroid pixel $(u,v)_j$.
From the list of blobs, the N highest-ranked candidates are identified and the associated events are aggregated and passed to separate asynchronous demodulation blocks.
If the blob corresponds to a smart visual beacon then, after synchronisation, the parity bits will be correct and the blob is confirmed as a beacon.
Non-beacon blobs are demoted to the bottom of the list of potential beacons after a period of incorrect parity bits (see \S~\ref{sec:End bit detection and synchronisation}).
The resulting available demodulation slot is then filled by the highest index blob in the detection list that is not already being demodulated.
In this way, the algorithm sorts through non-beacon blobs systematically, identifies beacon blobs and tracks active beacons in real time.
Successful beacons continue to be tracked until they lead to a sufficiently long sequence of incorrect parity bits, in which case they are demoted (see \S~\ref{sec:End bit detection and synchronisation}).

\begin{figure}[t!]
	\centering
	\begin{tabular}{c}
		\\
		\includegraphics[width=0.6\linewidth]{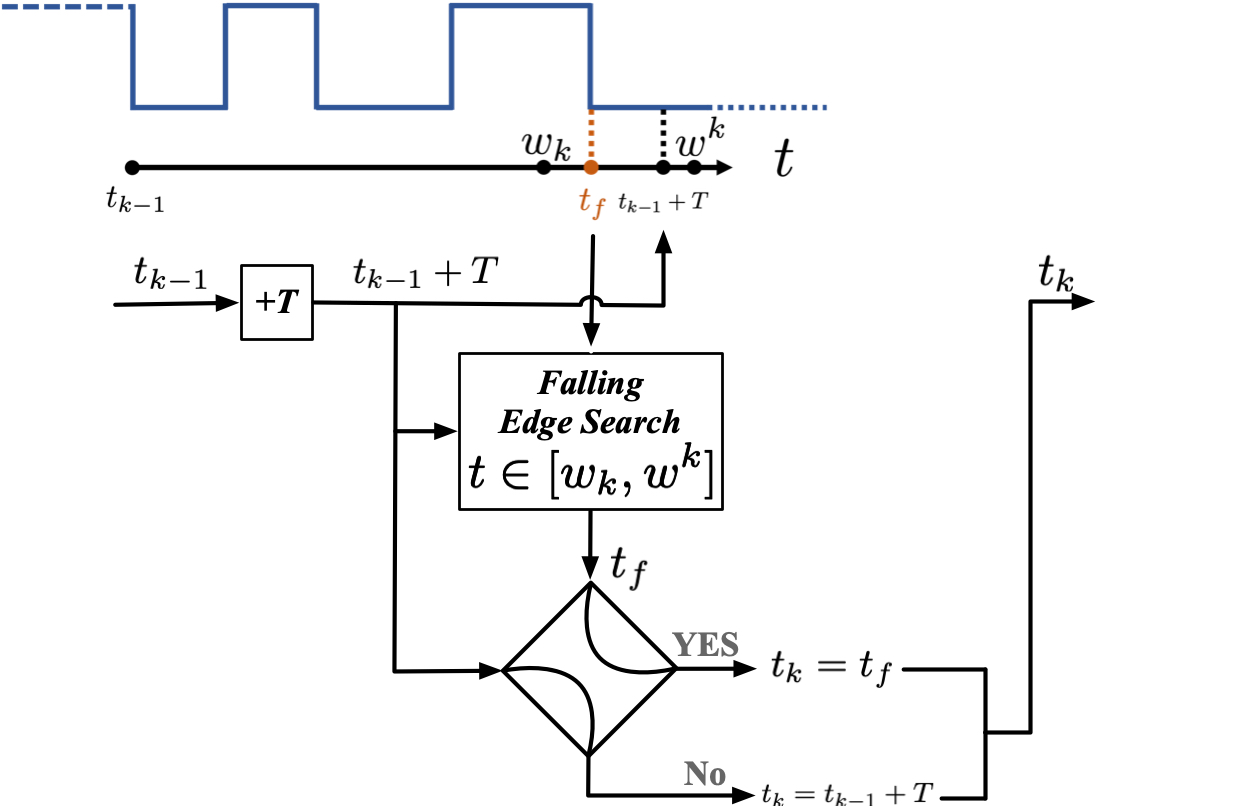}
	\end{tabular}
	\caption{End Bit Detection and Synchronisation.
		From the previous end bit $t_{k-1}$, our end bit detection algorithm finds the next end bit $t_k$ in the search window $t \in [w_k,w^k]$.
		If no falling edge is found, we choose the estimated time $t_{k-1} + T$ as the end bit and move to the next data packet.
		If synchronisation is lost, the algorithm will find a falling edge to the left of the present estimate and the new start bit estimate will slip to the left in the code.
		The new falling edge is not stable as the probability of a falling edge at this point in the code is only 25\%.
		Thus, the $t_f$ estimate will continue to slip to the left in the code and re-synchronise.
	}\label{fig:deconding}
\end{figure}

\subsubsection{Asynchronous high-pass filter and threshold}\label{sec:Asynchronous high-pass filter and threshold}
We integrate event data using the asynchronous event-based high-pass filter~\cite{scheerlinck2018continuous} which removes the low-frequency drift in event data while preserving the high-frequency variations triggered by the smart visual beacons.
The continuous-time ordinary differential equation (ODE) of the high-pass filter~\cite{scheerlinck2018continuous} is
\begin{align}
	\dot{\hat{L}}_{\bm{p}}(t) =  e_{\bm{p}}(t) - \alpha \hat{L}_{\bm{p}}(t),
	\label{eq:hf}
\end{align}
where $e_{\bm{p}}(t)$ is the event stream at pixel $\bm{p}$ and $\hat{L}_{\bm{p}}(t)$ is the corresponding continuous-time intensity estimate in log scale.
$\alpha$ is the pre-defined cutoff frequency which controls the decay rate of the intensity.
Note, however, that this ODE is asynchronously implemented and the solution is only computed when an event arrives.
The threshold is also only checked asynchronously when events occur minimising computational load.

We choose a cutoff frequency $\alpha$ that makes the integrated intensity decay to the reference intensity level quickly ensuring that there is no drift in the mean intensity.
This ensures that the next rising or falling edge is clearly visible.
In the experimental section we have used $\alpha = f/3$ where $f$ is the bit rate of the transmission signal ensuring the signal decays in approximately two bit periods (see the $\{0\}$ and $\{1\;1\}$ in Figure~\ref{fig:outdoor} b).

We define the high and low-level threshold values based on the camera sensitivity to positive and negative brightness changes.
Because event cameras usually have a positive bias that triggers more positive events than negative due to the undesirable junction leakage in the amplifier circuit or other noise~\cite{Wang19acra}, we choose a larger value as the high-level intensity threshold.
Figure~\ref{fig:outdoor}(b) shows the high-level and low-level threshold in red and blue horizontal lines.
The threshold trigger has hysteresis property built-in such that when the previous output is low, the input has to cross the high-level threshold before the output switches to high and vice versa.
This ensures that the output of the threshold module does not produce noisy switching output in between the two threshold (see dotted lines in Figure~\ref{fig:outdoor}(b) and Figure~\ref{fig:control voltage}).
In practice, the final binary signal is virtual and only the trigger times of rising and falling edges are retained and used in the packet decoding.

\subsubsection{End bit detection and synchronisation}\label{sec:End bit detection and synchronisation}
To detect the start bit of each character, we propose a novel synchronisation method that detects the falling edge (intensity levels jumps from high to low) triggered by the end bit $\{1\;1\}$ and the next start bit $\{0\}$ between two characters.
The method also automatically synchronises if the start bit is lost or when initialising.

In Figure~\ref{fig:deconding}, we define the time period $T = 1/f$ based on the data transmission frequency $f$.
From the previous end bit timestamp $t_{k-1}$, the next expected end bit timestamp is $t_k = t_{k-1} + T$.
Due to the noise in the hardware and the choice of threshold levels, the position of falling edge timestamp might shift slightly.
Therefore, we search for the first falling edge from the right in a search window
\begin{align}
	t & \in [w_k, w^k],\\
	w_k &= t_{k-1} + T + \Delta t,\\
	w^k &= t_{k-1} + T - 5\Delta t,
	\label{eq:search window}
\end{align}
where the $\Delta t$ is the expected time period for half a bit.
For the UART encoding used in this paper $\Delta t = T/22$.
The search window has a wider range to the left ($5\Delta t$) since the end bit $\{1\;1\}$ contains two bits but the start bit $\{0\}$ is only 1 bit.
In particular, if $ t_{k-1} + T $ lies within $\Delta t$ of the true start bit falling edge $t_f$ there cannot be another falling edge to the right of $t_f$ in the search window.
In contrast, if the signal loses synchronisation and drifts to the left, the final bit transition from message to parity bit becomes visible in the left hand edge of the search window.
With probability 25\% (corresponding to a $\{1\;0\}$ sequence) this will be a falling edge and the new $t_f$ will be shifted to the left in the code.
This property plays an important role in the synchronisation module which is explained below.

If no falling edge is found in the search window, we choose $t_k = t_{k-1} + T$ and the next character is decoded based on this choice.
In this way, the falling edge can be missed without compromising the synchronisation of the data stream.

If the synchronisation is lost, or during initialisation where $t_{k-1}$ is chosen as the first falling edge encountered, then the falling edge $t_f$ found in the search window no longer corresponds to the start bit.
In this case the probability that a falling edge occurs at, or close to, $t_{k-1} + T$ is at best 25\%, since this requires a $\{1\;0\}$ sequence of bits to occur exactly in this slot.
If there is no falling edge at $t_{k-1} + T$ then, due to the search window design, there is a high chance that another falling edge will be detected to the left.
This causes the next start bit time estimate $t_k$ to occur at least one place earlier in the code packet.
After at most 10 such slips the new falling edge found should be the true start bit, and from then on the signal should remain synchronised.

An incorrect choice of start bit will result in an incorrect parity bit roughly 50\% of the time.
It is clear that if the parity bit is incorrect on sufficiently many characters from a long decoded sequence, the chance that the signal comes from a beacon is low.
The full details of the probability analysis associated with accepting or rejecting a beacon are complicated and beyond the present development.
However, the authors found that a sequence of 15 characters that terminates without at least 4 of the last 5 parity bits correct is a robust criterion to reject false beacons.
In practice, most signals synchronise within 5 characters.

\subsubsection{Discretisation and decoding}\label{sec:Discretisation and decoding}
We discretise each data packet into an 11-bit binary number by dividing the time period between the start and end bits into 11 slots.
The values of the bits are computed by working forward through the time offsets of the identified rising and falling edges in the sequence.
The parity bit is checked first and if correct the character bits are decoded from the ASCII code.

\section{EXPERIMENT}
\subsection{Datasets}
\begin{figure}[t!]
	\centering
	\begin{tabular}{c c}
		\\	\includegraphics[width=0.33\linewidth]{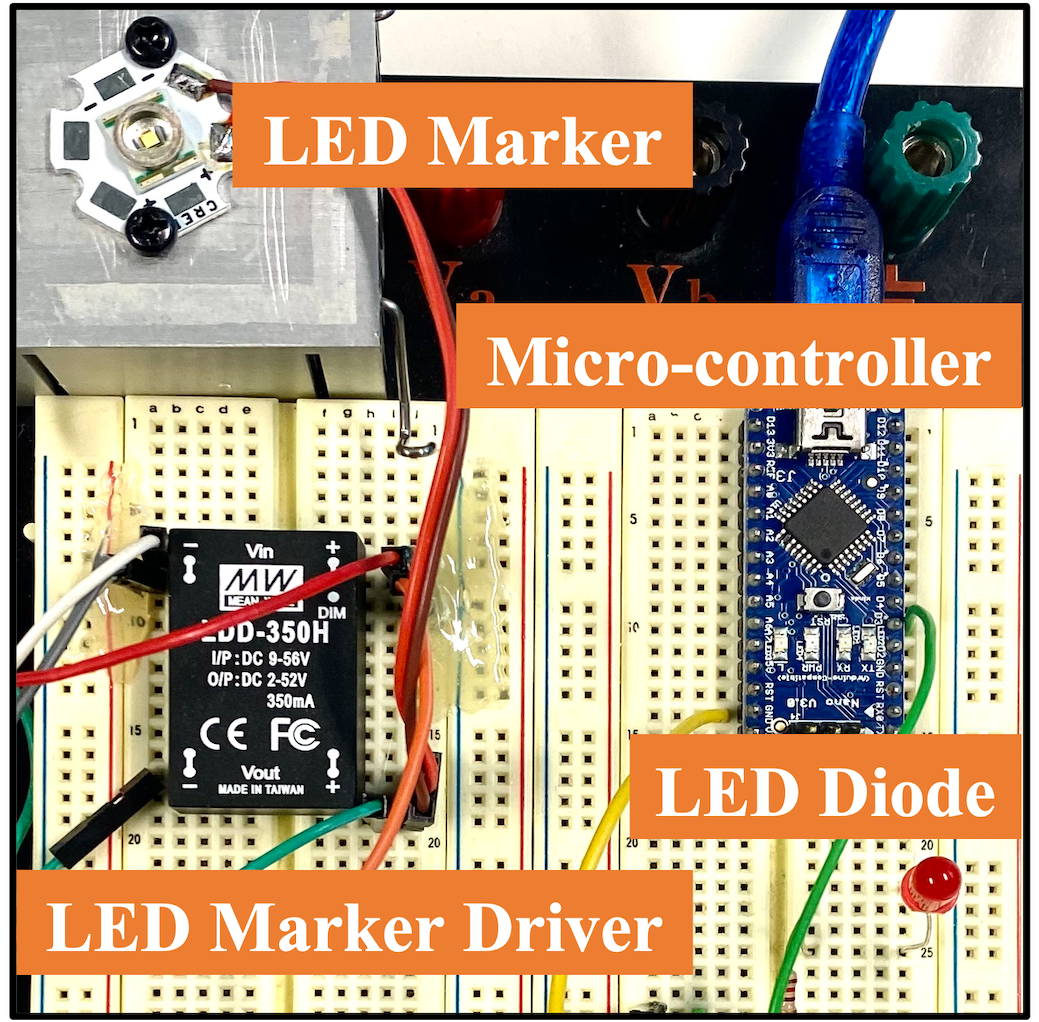} &
		\includegraphics[width=0.33\linewidth]{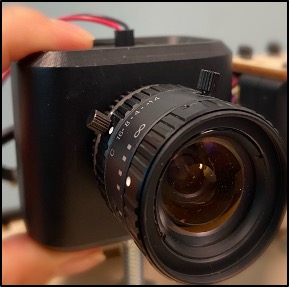} \\
		(a) LED Circuit  & (b) Event Camera
	\end{tabular}
	\caption{\label{fig:circuit and camera}
		Our Event-based Smart Visual Beacon Communication System.
	}
\end{figure}

The prototyped LED circuit is demonstrated in Figure~\ref{fig:circuit and camera} (a).
We evaluate the communication system using a dim 5 mm LED (14 mA) that is directly current controlled by a resistor, and a brighter LED marker (CREE XR-E Q5) controlled by a constant current driver (350 mA).
We use a \textit{Gen3 Prophesee} event camera (VGA, $640\times480$ pixels) to capture the blinking LED lights.
For the indoor experiment, the event camera is mounted with its original lens, a \textit{computar} f/1.4, 8 mm lens.
For a long-distance communication in the outdoor experiment, we replace the original lens with a \textit{Nikon} lens, with f/4.5-5.6 70 mm, and field of view of $22^{\circ}50'$.

The message we transmitted is the Sonnet 18: ``Shall I compare thee to a summer's day?" by William Shakespeare~\cite{Sonnet}, which includes 618 characters.
In total, 6798 bits are transmitted for each sonnet.
We transmit the Sonnet twice, with a 16-bit high-level signal between two messages, and let the algorithm automatically detect the start timestamp of the second Sonnet.

\subsection{Control voltage and event camera}
\begin{figure}[t!]
	\centering
	\begin{tabular}{c}
		\\	\includegraphics[width=0.49\linewidth]{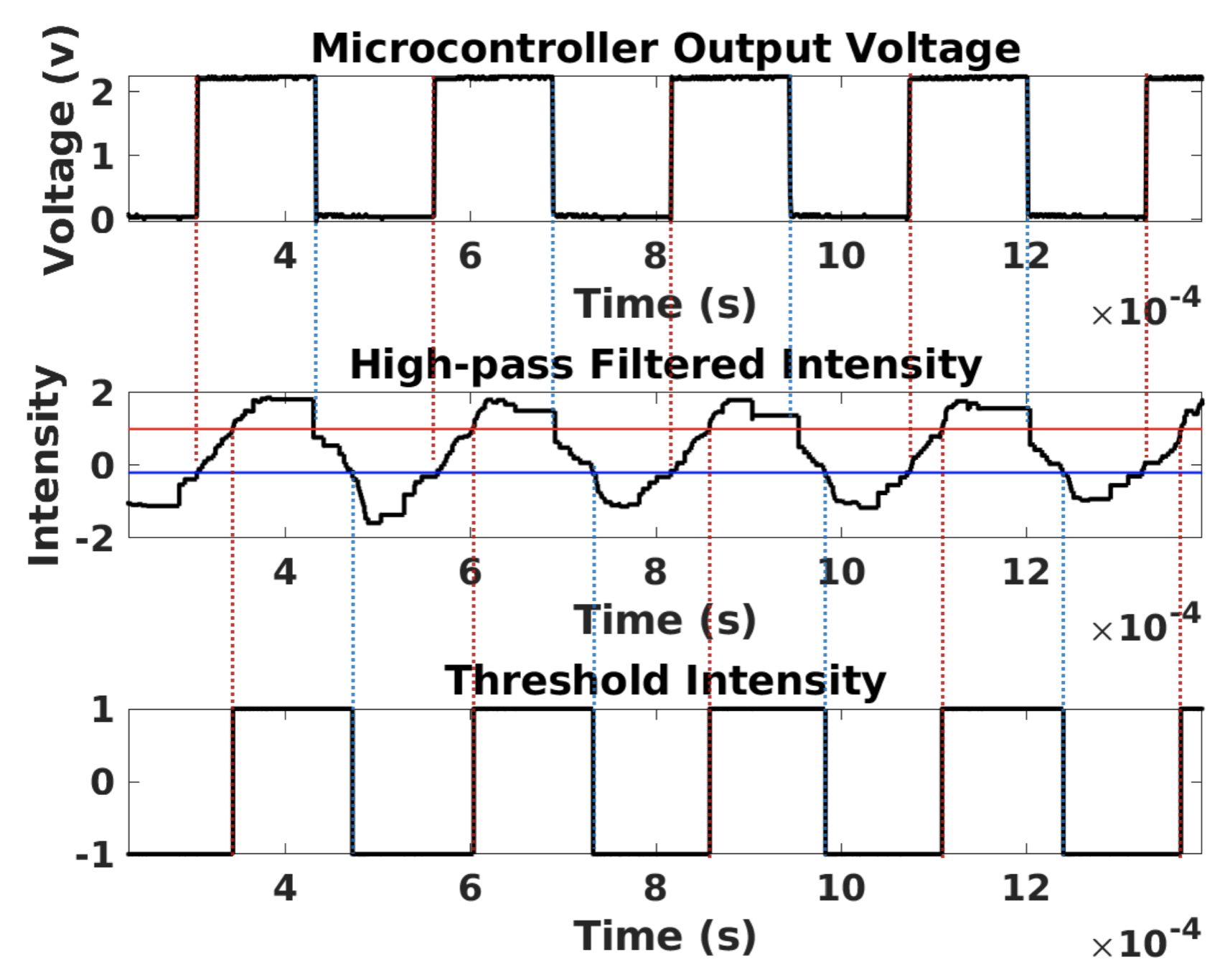}
		\\
	\end{tabular}
	\caption{
		Control Voltage from a Microcontroller and the High-pass Filtered~\cite{scheerlinck2018continuous} Event Data Comparison.
		The microcontroller provides a 4 kHz square wave control voltage to the LED lights.
		The event camera is able to capture the high-frequency blinking signal with high accuracy.
	}\label{fig:control voltage}
\end{figure}

We first evaluate the ability of an event camera to capture the high-frequency flashing LED lights by comparing the controlled voltage from a microcontroller and event data captured by an event camera.
We use a microcontroller to generate a square wave control voltage at 4 kHz and directly measure the output voltage of the microcontroller using an oscilloscope.
The LED is directly connected to the microcontroller TTL signal to minimize signal non-linearity due to power electronics but this leads to very dim LED light.
We use an event camera to capture the LED blinking signal at a distance of 30 cm.
We apply the high-pass filter~\cite{scheerlinck2018continuous} to the asynchronous event data to generate the continuous-time intensity and then use two high-level and low-level thresholds to binarise the signal (see the first row and second row in Figure~\ref{fig:control voltage}).
The high and low-level thresholds are shown in red and blue lines respectively.
The microcontroller output voltage is plotted in the first row of Figure~\ref{fig:control voltage}.
It shows that the integrated event data matches the control voltage in shape which demonstrates the ability of an event camera to capture the high-frequency LED blinking with high accuracy.

\begin{figure}[t!]
	\centering
		\begin{tabular}{cc}
			\\	\includegraphics[width=0.48\linewidth]{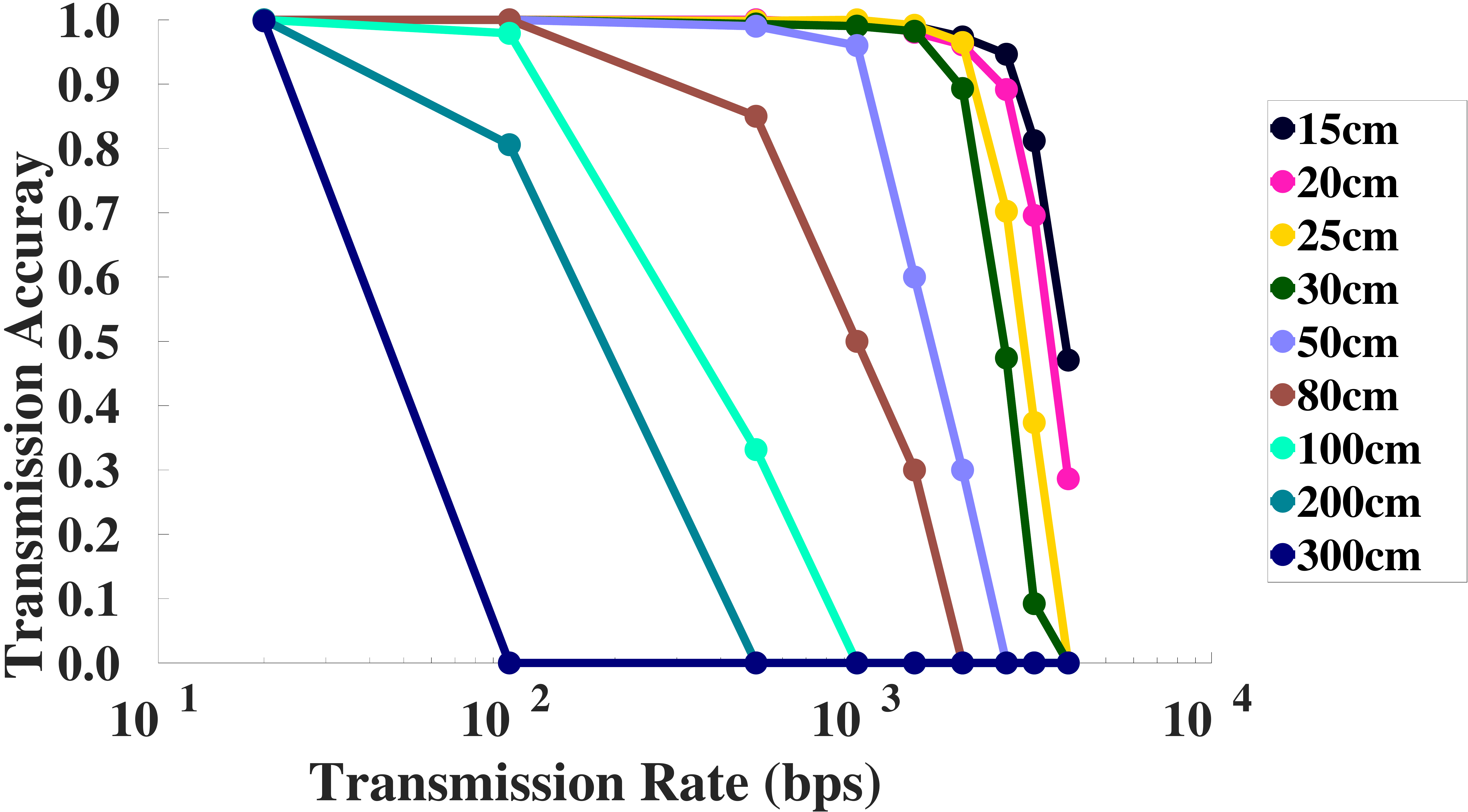} & \includegraphics[width=0.48\linewidth]{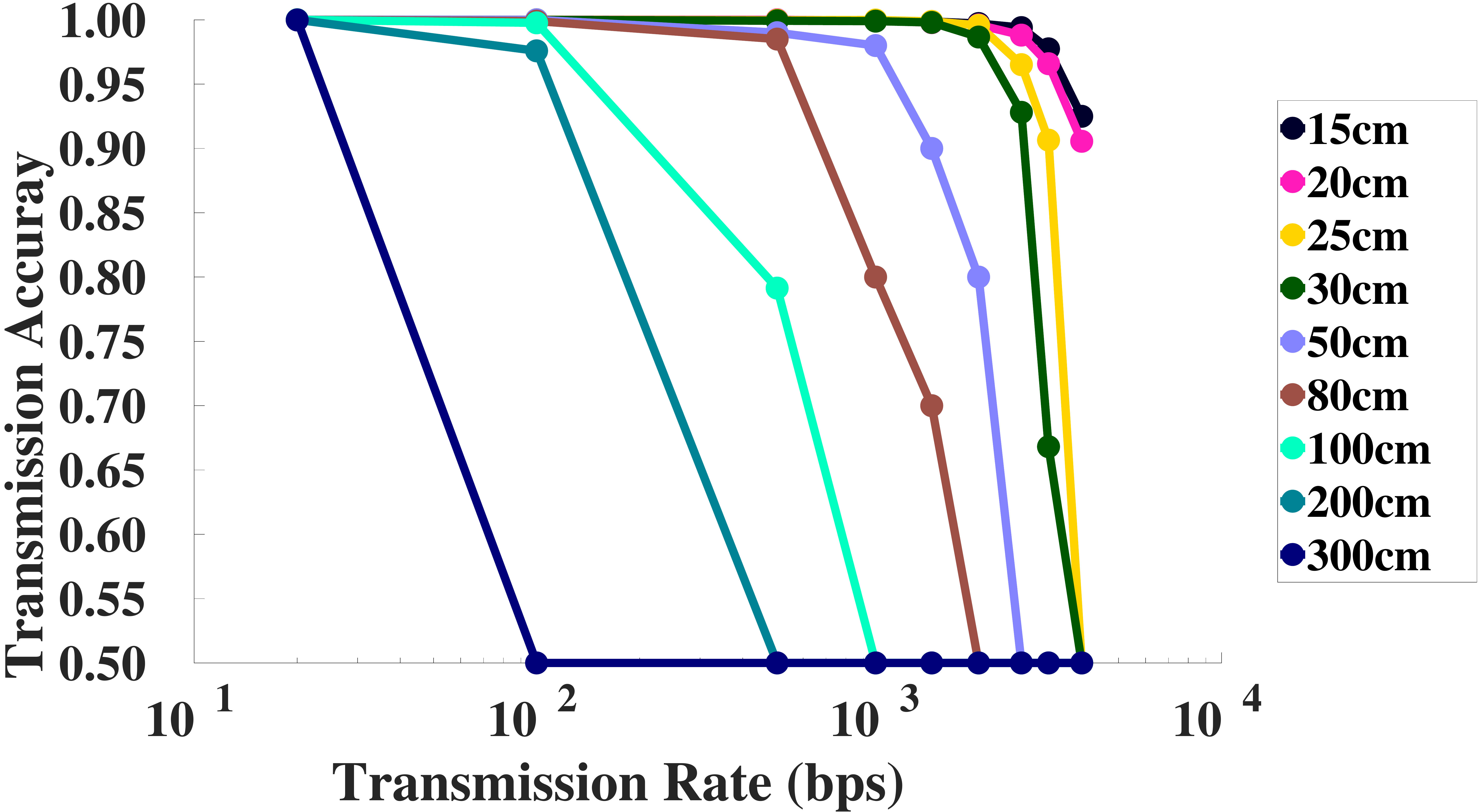} \\
			(a) Message Accuracy Rate & (b) Bit Accuracy Rate \\
		\end{tabular}
	\caption{Indoor Experiment.
		We evaluate the relationship of the transmission accuracy defined as message accuracy rate in (a) and bit accuracy rate in (b) versus LED transmission rate.
		A range of communication distances were tested for a dim LED (14 mA) in an indoor setting.
		The results clearly demonstrate the degradation of message quality with increasing distance and transmission rate.
		Note that it is still possible to transmit information at longer distance if the frequency is chosen sufficiently small as demonstrated by the 100\% transmission accuracy at 3 meters for 20 bps transmission rate.
	}\label{fig:indoor}
	\vspace{-2mm}
\end{figure}

\subsection{Main results}
We quantitatively evaluate the communication system using two metrics: the message error rate (MER) and the bit error rate (BER).
The message error rate is the number of wrongly decoded characters divided by the total number of characters, and the BER is the number of bit errors divided by the total number of transmitted bits in the whole dataset.
Extreme long-distance communication or high transmission rate might lead to noisy event data.
Hence, we set the MER to 1 and BER to 0.5 for datasets that cannot accurately identify the start timestamp of the transmitted message.
We define the accuracy as the complement of the error rate.
So the transmission message accuracy rate is (1 - MER) and the bit accuracy rate is (1 - BER).

\subsubsection{Indoor experiment}
In Figure~\ref{fig:indoor} we demonstrate the relationship between the LED transmission rate and the communication distance in an indoor daylight environment.
It shows that a longer communication distance requires a lower transmitted rate.
In a distance shorter than 50 cm, our smart visual beacon system is able to communicate more than 1 kbps signal with more than 96\% message accuracy rate and 98\% bit accuracy rate.
Even with a dim LED and a small machine vision lens, our system achieves 100\% transmission accuracy in a communication distance of 3 meters with a lower data transmission rate of 20 bps.
The MER only measures the accuracy of the decoded characters (8-bit ASCII), which requires 100\% accuracy for all 8-bit ASCII code to be successfully decoded but does not consider the start and end bits.
The BER reflects the accuracy of all binary bits.
Therefore, when the signal-to-noise level increases, the message accuracy rate decreases to 0 quickly, but the bit accuracy rate decreases to 0.5 with a relatively slower speed.
Notably, since the LED is dim and the curve is highly dependent on the environmental lighting conditions, we keep all experiments in similar indoor daylight lighting conditions.
The curve would shift to the right for an experiment in a darker environment or when using a brighter LED, and the communication system would perform better.

\subsubsection{Outdoor experiment}
\begin{figure}[t!]
	\centering
	\begin{tabular}{c}
		\includegraphics[width=0.49\linewidth]{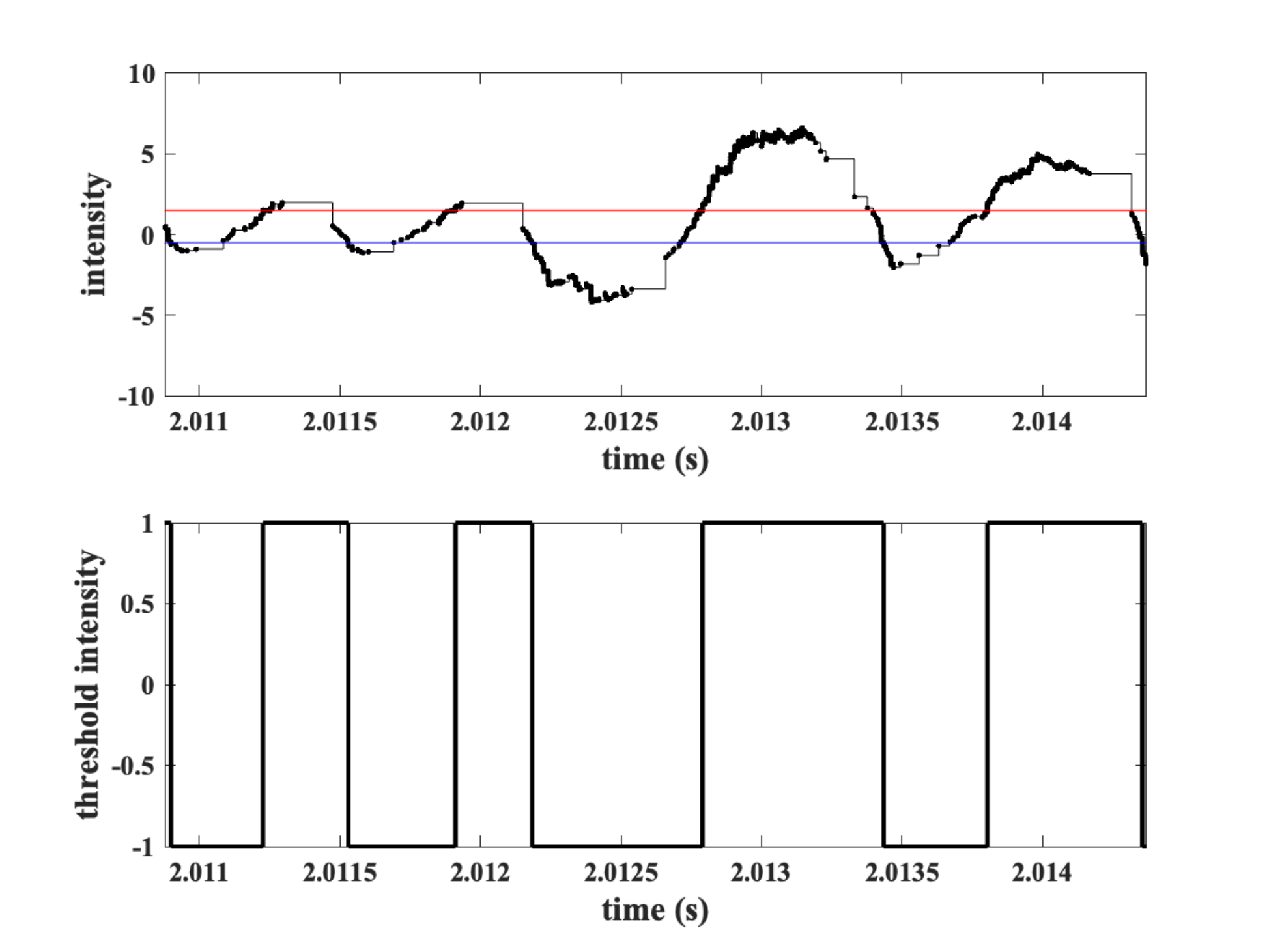} \\
	\end{tabular}
	\caption{		
		Outdoor Experiment Data. Example of 3125 bps transmission at 100 m. Blue and red line show the low and high-level contrast threshold respectively.
		Note the poor signal response at high frequency primarily due to the LED power supply. 
	}
	\label{fig:outdoor3000bps}
\end{figure}

\begin{table}[]
	\centering
	\caption{Message Accuracy Rate and Bit Accuracy Rate for the Outdoor Experiment.}
	\begin{tabular}{ccccc}
		\toprule \midrule[0.3pt]
		& \multicolumn{2}{c}{Message Accuracy Rate}          & \multicolumn{2}{c}{Bit Accuracy Rate}    \\    \cmidrule{2-5}
		& \multicolumn{1}{c}{30 m} & 100 m & \multicolumn{1}{c}{30 m} & 100 m \\ \midrule[0.8pt]
		500 bps  & \multicolumn{1}{c}{1}     &  1     &  \multicolumn{1}{c}{1}   &   1    \\  \hline
		3125 bps & \multicolumn{1}{c}{0.877}    &    0.435   & \multicolumn{1}{c}{0.987}     &  0.906   \\ \midrule[0.3pt]  \bottomrule
	\end{tabular}\label{tab: outdoor}
\end{table}

We evaluate the smart visual beacon communication with a transmission rate of 500 bps and 3125 bps in several outdoor daylight environments to demonstrate long-distance transmission capability.
For the outdoor experiments, a separate power circuit for the LED light was used to ensure sufficient brightness for an outdoor environment.
The power electronics, however, introduced voltage slew rate limitations that led in turn to reduced signal response at high frequency.
We tested the outdoor system at 500 and 3125 bps over 7 m, 14 m, 20 m, 30 m, 50 m, 100 m.
Figure~\ref{fig:outdoor} (a) demonstrates the two different outdoor experiment environments.
Due to space limitations we have included only the results at 30 m and 100 m.

The complex background motion of trees, people and cars does not affect the transmission accuracy.
Figure~\ref{fig:outdoor} shows the response of the system for the lower 500 bps transmission rate while Figure~\ref{fig:outdoor3000bps} shows the response for the higher transmission rate. 
The first row of the figure in both cases shows the high-pass filtered received signals and below shows the decoded bit sequence. 
The experiment demonstrates 100-meter error-free transmission at 500 bps, superior to existing systems with a similar cost and complexity~\cite{saeed2019optical}.
For higher bit rates, we expect to see degradation in the signal transmission 
(Table~\ref{tab: outdoor}). 
The message accuracy rate degrades much faster than the bit rate degradation as expected. 
Degradation of the outdoor signal is primarily due to reduced intensity of the received signal and consequent failure to trigger the thresholds. 
This is partly due to distance and partly due to the voltage slew rate limitations of the LED power electronics.
A dedicated power electronics system designed to maximise the frequency response of the LED would significantly improve outdoor performance. 
However, any system would eventually be limited by the refractory response of the camera as seen in the indoor experimental results.

\section{CONCLUSION}
In this paper, we propose a novel architecture for the event-based smart visual beacons by exploiting the strengths of event cameras.
We prototype the communications channel of the smart visual beacons and evaluate the performance in both indoor and outdoor environments.
In indoor environments, we achieve up to 4 kbps data rate, while for outdoor we achieve 100-meter 500 bps error-free communication in daylight conditions, demonstrating the potential of this technology for distributed robot systems in real-world applications.

\section{ACKNOWLEDGMENTS}
The authors would like to thank Yicong Hong for helping data collection, and thank PROPHESEE for providing the event camera that was used in the work.

\clearpage
\bibliographystyle{ieee_fullname}
\bibliography{template_arxiv}

\end{document}